\documentclass[conference]{IEEEtran}

\IEEEoverridecommandlockouts
\usepackage{cite}
\usepackage{amsmath,amssymb,amsfonts}
\usepackage{algorithmic}
\usepackage{graphicx}
\usepackage{textcomp}
\usepackage{xcolor}
\usepackage{subfigure}
\usepackage{siunitx}

\usepackage{color}
\usepackage{wrapfig}
\usepackage{hyphenat}
\usepackage{url}
\usepackage[affil-it]{authblk}

\title{Dynamic Autonomous Surface Vehicle Control and Applications in Environmental Monitoring
\thanks{The authors would like to thank the National Science Foundation for its support (NSF 1513203). The authors would like to acknowledge the help of the College of Engineering and Computing, University of South Carolina for their support.}
}
\author{Nare Karapetyan, Jason Moulton, and Ioannis Rekleitis}
\affil{Computer Science and Engineering Department, University of South Carolina }
\date{}

\begin{document}
 \maketitle
 \begin{abstract}
This paper addresses the problem of robotic operations in the presence of adversarial forces. We presents a complete framework for survey operations: waypoint generation, modelling of forces and tuning the control. In many applications of environmental monitoring, search and exploration, and bathymetric mapping, the vehicle has to traverse in straight lines parallel to each other, ensuring there are no gaps and no redundant coverage. During operations with an Autonomous Surface Vehicle (ASV) however, the presence of wind and/or currents produces external forces acting on the vehicle which quite often divert it from its intended path. Similar issues have been encountered during aerial or underwater operations. By measuring these phenomena, wind and current, and modelling their impact on the vessel, actions can be taken to alleviate their effect and ensure the correct trajectory is followed.
\end{abstract}
 \begin{IEEEkeywords} autonomous surface vehicles, control, coverage, autonomous systems
\end{IEEEkeywords}

\section{Introduction}

\begin{figure}[ht]
\centering
\includegraphics[width=0.4\textwidth]{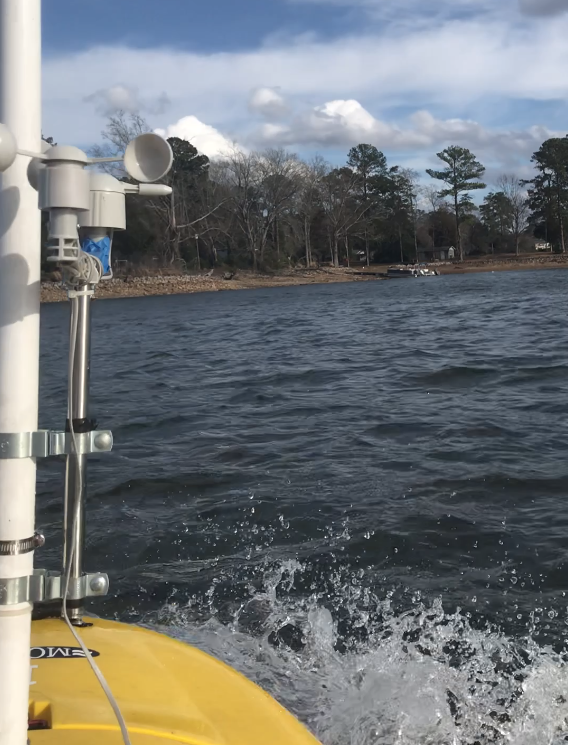}
\caption{ASV platform collecting wind data.}
\label{fig:beauty} 
\end{figure}

\begin{figure*}[ht]
		\centering
		\leavevmode
		\begin{tabular}{cc}
		   \subfigure[]{\includegraphics[width=0.49\textwidth]{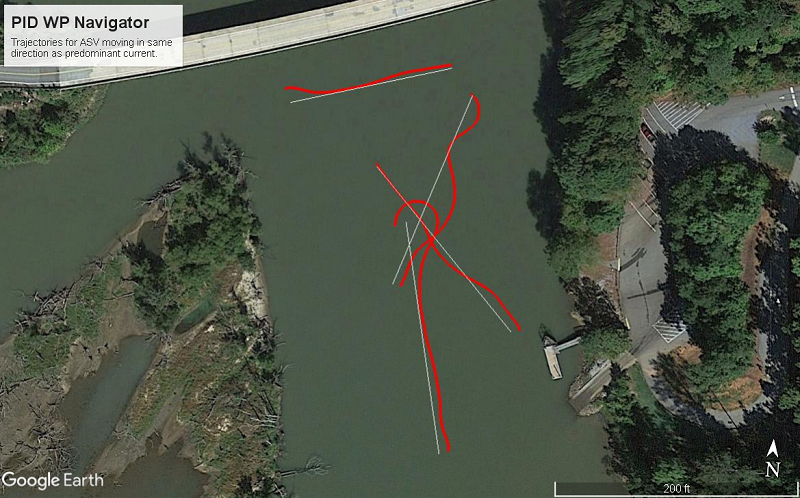}\label{fig:p1}}&
			\subfigure[]{\includegraphics[width=0.49\textwidth]{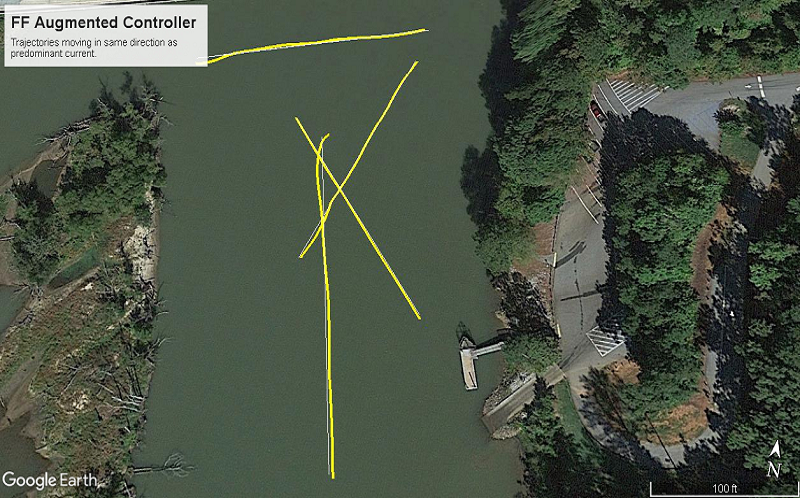}\label{fig:p2}}
		\end{tabular}
			\caption{The trajectory of ASV way\hyp point navigator (a) only using the PID controller on Pixhawk; (b) with augmented waypoint navigation.
			\label{fig:augmented_results}}
	\end{figure*}

When operating on dynamic aquatic environments the precise and efficient control is important especially for small vehicles (see Figure \ref{fig:beauty}). While wind or small changes in current might not drastically affect the trajectory of large ships, it adds significant noise in the execution of mission plans of small Autonomous Surface Vehicles (ASVs); (see Figure \ref{fig:augmented_results} (a)). In order to achieve optimal path planning for the coverage of an area, an accurate point to point navigation is a key component. In this work we present a complete pipeline of achieving adaptive control which takes into account the error in displacement caused by the forces such as water currents and wind in order to increase the accuracy.

In order to execute any decisions in dynamically changing environments we collected extensive data along different trajectories during different environmental conditions. In particular water current, wind sensor data, along with the boat's GPS coordinates and heading information were recorded. The control can be adjusted in two modes: based on predictive values of water and current or for the current readings of sensors. In the first scenario a model of the environment is used represented by the Gaussian Processe. In the second scenario each measurement is processed as it is received during the execution. After this step the strategy is the same for both scenarios: the speed and orientation of the ASV is used to determine the absolute values of each measurement which is used as an input for the linear regression to predict the effect of the forces on the speed and direction of ASV.
 While the target way-point is not reached, an intermediate way-point is calculated based on the effect values. The speed is also adjusted based on the predicted error. Finally, the ASV is sent to the newly calculated way-point. When the new target position is processed by the navigation controller, it results in a smoother and more accurate path, following the planned trajectory. And finally the enhanced control is used to improve accuracy of different coverage techniques proposed for lake and river monitoring operations.

\begin{figure}[ht]
\centering
\includegraphics[width=0.48\textwidth]{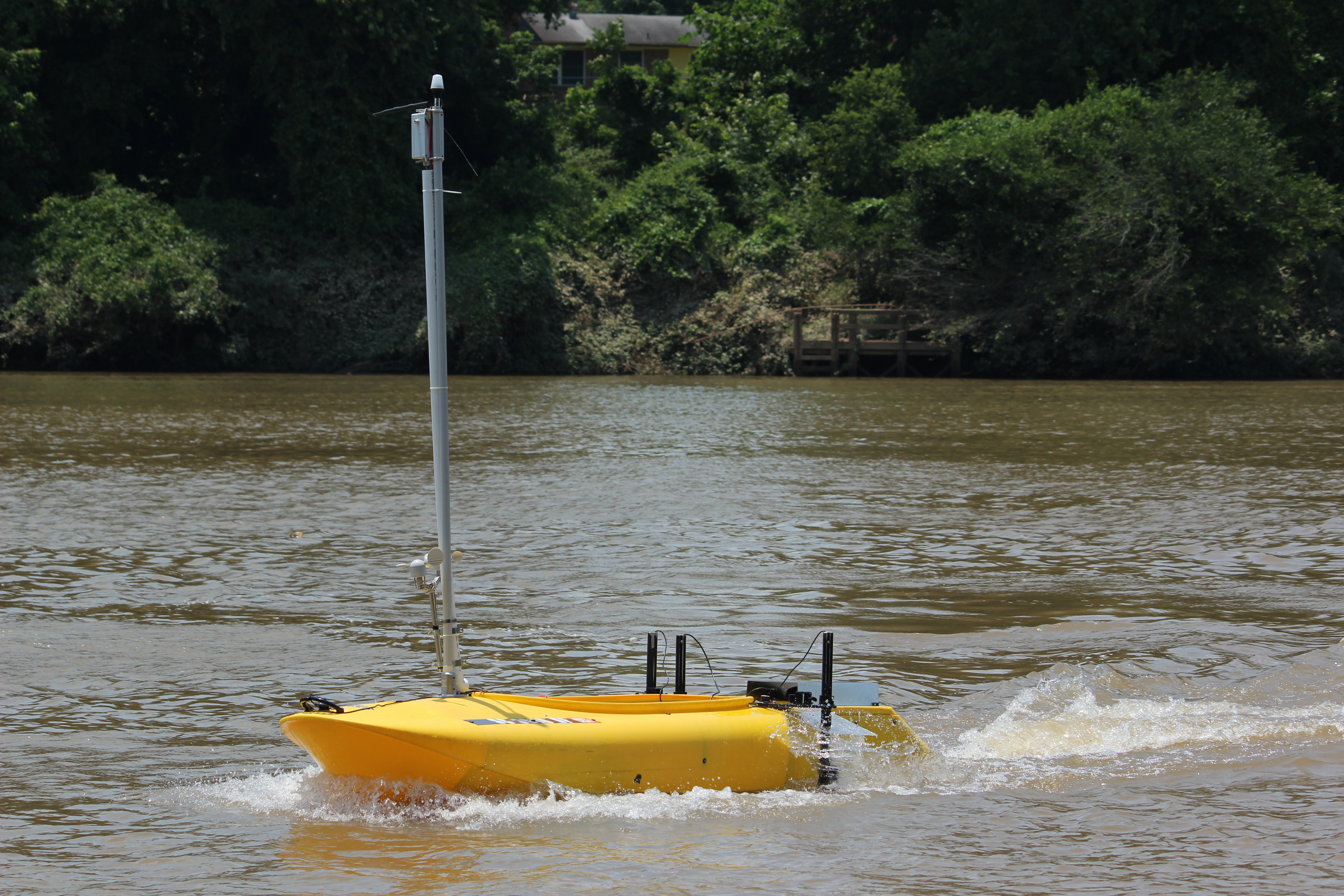}
\caption{ASV platform used for deployment.}
\label{fig:ASV} 
\end{figure}

All deployments of the proposed method were performed on a Jetyak Autonomous Surface Vehicle (ASV) designed in AFRL \footnote{https://afrl.cse.sc.edu/afrl/resources/JetyakWiki/} \cite{moulton2018autonomous} (see Figure \ref{fig:ASV}). The boats are equipped with a Pixhawk controller used for waypoint navigation, a Raspberry  Pi computer that runs the Robot Operating System (ROS) to log sensor data and perform the MAVLink based navigation. The main sensors used on the vehicle are an NMEA 0183 depth sonar, a Sparkfun anemometer for wind, and the Ray Marine ST 800 paddle wheel speed sensors for current measurements. For analog to digital conversion of current and wind sensors the drivers are provided by ArduinoMega and Wearhershield micro-controllers. The experiments were performed both on lake Murray and the Congaree river with different possible patterns to demonstrate the effect of the forces on the trajectories. 

\begin{figure*}[h]
		\centering
		\leavevmode
		\begin{tabular}{cc}
		   \includegraphics[width=0.5\textwidth]{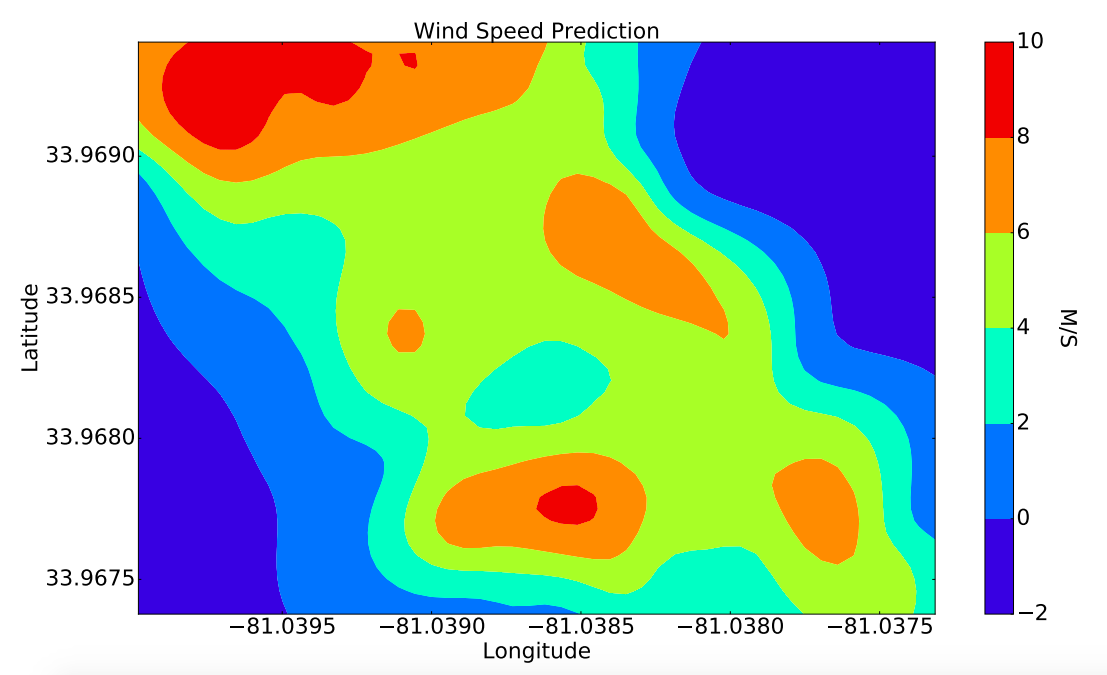}\label{fig:p1}&
		   \includegraphics[width=0.5\textwidth]{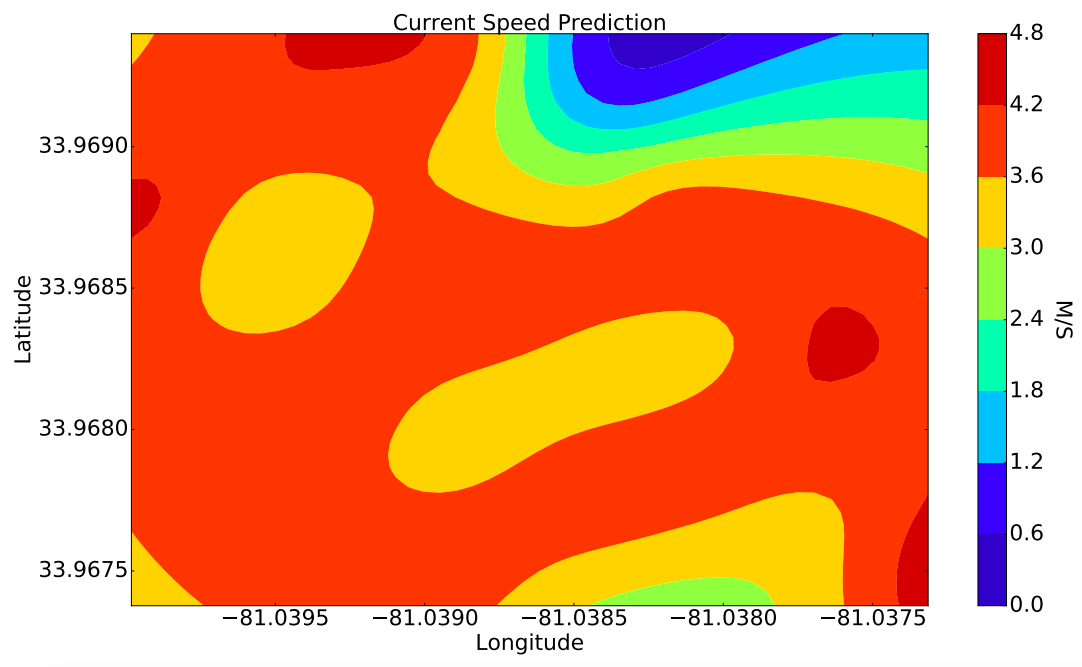}\label{fig:p2}
		\end{tabular}
			\caption{Force Maps of Congaree River: (a) wind speed map, (b)water current speed map.
			\label{fig:gaussian}}
	\end{figure*}

\section{Related Work}

Different research groups have been working on development of autonomous surface vehicles for environmental monitoring \cite{whoiMokai2014, mahacek2008development, GirdharIROS2011, fraga2014squirtle, curcio2005scout}. Similar to the design presented by Kimball et al. \cite{whoiMokai2014} the ASV system proposed by our lab is based on the gas powered Mokai kayak \cite{moulton2018autonomous}. 

	\begin{figure*}[ht]
		\centering      
		\includegraphics[width=0.8\textwidth]{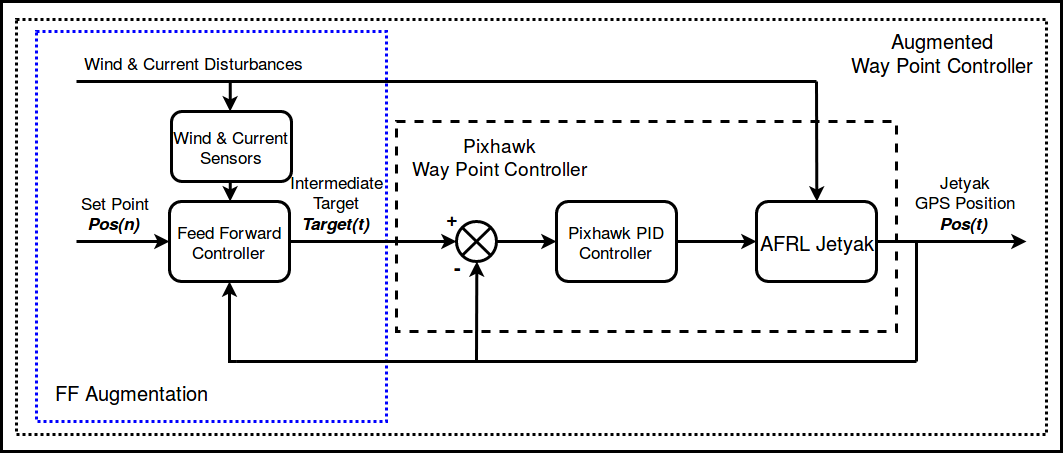}
		\caption{The way\hyp point navigation PID controller used in the Pixhawk PX4 augmented by our intermediate way\hyp point offset generator.}
		\label{fig:ff_con}
	\end{figure*}

The literature has in depth studies on the control of different vehicles and namely surface vehicles under environmental forces. In particular Rasal has proposed a method for overcoming disturbances by measuring the IMU error\cite{sclara}. Digital feed-forward tracking controller has been designed by Tsu-Chin Tsao \cite{tsao1994optimal} that resembles our approach but in contrast does not take into account environmental forces. Some control approaches have been proposed also based on Neural Networks, such as work by Pan et al. \cite{pan2013efficient}. Even though the proposed NN approach is capable of tracking the desired trajectory with good control performance by on-line learning, the experimental setup of the work is only based on the simulation and might not be appropriate for reactive onboard control. 

\begin{figure}[ht]
\centering      
    \includegraphics[width=0.45\textwidth]{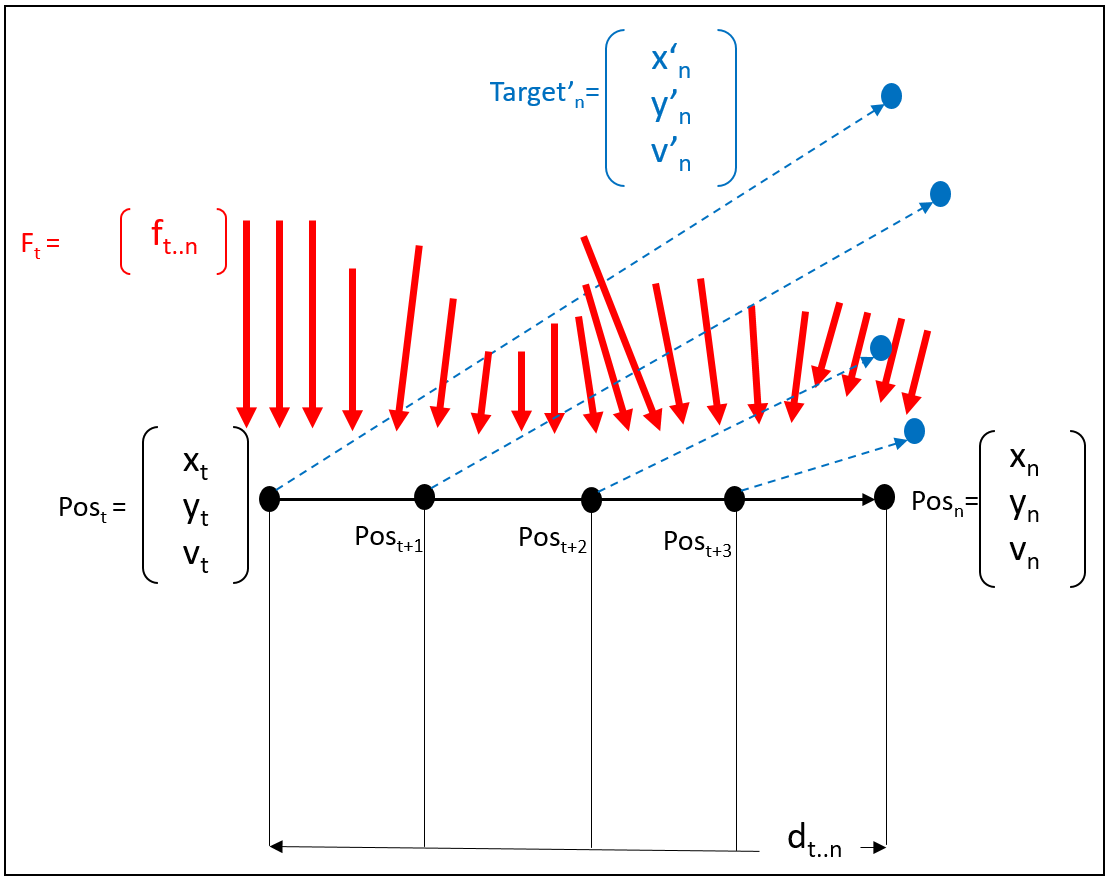}
	\caption{High\hyp level illustration of way\hyp point navigation augmentation method.}
	\label{fig:carrot}
\end{figure}

	\begin{figure*}[h]
		\centering
		\leavevmode
		\begin{tabular}{cccc}
		   \subfigure[]{\includegraphics[height=1.39in]{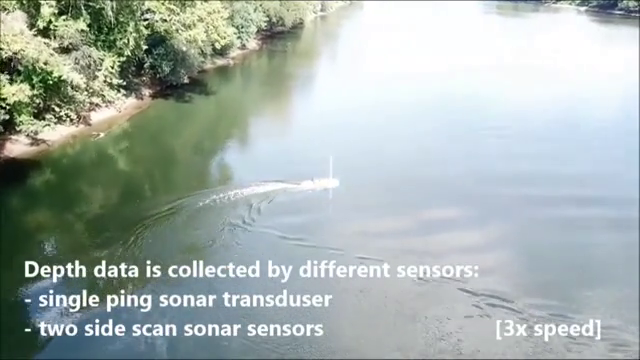}\label{fig:p1}}&
		   \subfigure[]{\includegraphics[height=1.39in]{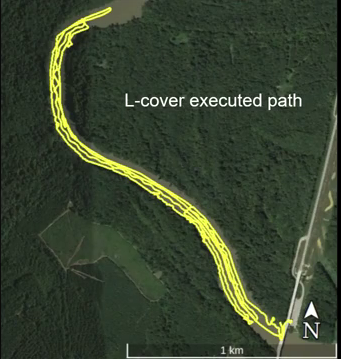}\label{fig:p1}}&
		   \subfigure[]{\includegraphics[height=1.39in]{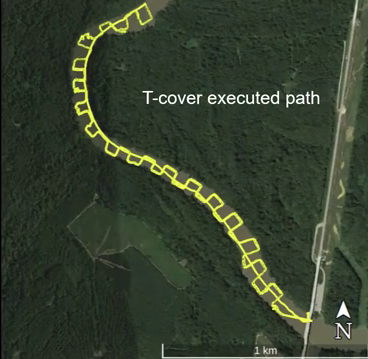}\label{fig:p1}}&
			\subfigure[]{\includegraphics[height=1.39in]{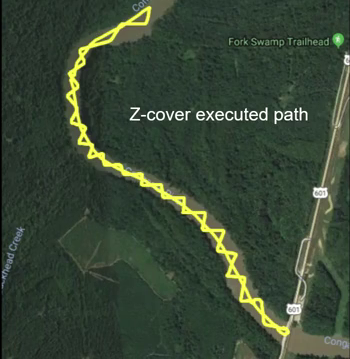}\label{fig:p2}}
		\end{tabular}
			\caption{Different Coverage patterns (a) an ASV performing coverage on a Congaree River, SC, USA; (b) Longitudinal L-cover pattern; (c) Transverse T-cover pattern; (d) Zig and zag Z-cover pattern.
			\label{fig:coverages}}
	\end{figure*}

 Hsieh et al. \cite{hsieh2012robotic} have tackled the problem of mapping the effects of current phenomena using tracking control strategy of collaborative underwater robots. Another work proposed by Huynh et al. presents a path planning method for minimizing the energy consumption of an autonomous underwater vehicle (AUV) \cite{huynh2015predictive}. In this work different ocean disturbances are studied, which are assumed are not exceeding the capabilities of the AUV. Even though these works, addressing collection and modeling of dynamic environmental properties, they do not provide the time and space considerations that impact a lightweight, small-scale ASV. To implement a future adaptive controls and online path planning techniques for surface vehicles a reliable map of environmental forces is required and in this work we generate those with Gaussian Processes \cite{MoultonISER2018}.

When solving the coverage problem the assumption is that the vehicles control is smooth and it is not considered as an efficiency constrain. Thus most of the coverage algorithms ended up being deployed as a waypoint mission. Some of the notable works in area coverage are summarised in a survey paper by Galceran and Carreras \cite{galceran2013survey}. This work is an attempt to combine control with coverage path planning algorithms to accommodate efficient and smooth waypoint mission execution.

\section{Proposed Approach}

The proposed framework allows to consider path planning strategies, use the onboard sensors to predict the effect of the force on the navigation correctness and perform correction by introducing intermediate waypoints. The proposed approach is using an ASV equipped with waypoint navigation, wind and water current sensing capabilities. The framework has the following components: modelling the environmental forces, performing prediction of the affect of the the forces on the navigation, augmentation of waypoints or a feed-forward PID controller, and finally the offline generation of waypoints for complete or partial coverage. 

\paragraph{Modelling Environmental Forces}
When performing waypoint navigation it is sometimes desirable to perform corrections beforehand rather that reactively by current measurements of sensors. Collecting wind and water current data, we build a map of the external forces speed and direction using Gaussian Processes with Matern 3/2 kernel \cite{MoultonISER2018} (see Figure \ref{fig:gaussian}). 

\paragraph{Prediction of the Force Affect}
For reactive response the direct measurements are taken from the sensors. First a comprehensive data is collected to train a linear model for predicting the displacement effect from the current sensor readings. The displacement effect is calculated as x and y components of the error distance from desired location and the actual location of the boat. The trained linear model is used online for prediction of this effect.

\paragraph{Augmentation of waypoints}

When the linear model is available either using the current sensor readings or the predicted values from the Gaussian Process we perform a waypoint augmentation. By changing the target global pose based on the measurements and effects of the external forces an intermediate waypoint is generated for smoothing the trajectory. This represents a version of a feed-forward controller (see Figure \ref{fig:ff_con}) where instead of changing the velocity and heading, the corrections are performed by the introduction of waypoints. Figure \ref{fig:carrot} illustrated the main idea of this stage: black solid line and position points denote the desired path that should be maintained. Blue arrows represent the wind and current force vector acting on the ASV. Red points and arrows represent the intermediate way\hyp points provided to the Pixhawk navigator and their associated target headings.

\paragraph{Offline Waypoint Generation}

To complete the proposed framework, effective monitoring techniques are required. Our system provides different coverage methods for different environments and different survey operations. We have developed efficient multi-robot coverage algorithms for lakes that take into account the rotation angle constraint that the ASVs have \cite{karapetyan2018multi}. There are two approaches used for tackling multi-robot coverage problem in this work. First approach finds a single robot coverage path planning and then splits that between the multiple-robots. The second approach splits the area between robots by the size of the coverage region and the distance from the starting point, and then finds for each robot the single optimal path. 
In addition we have developed different coverage patterns designed for river surveying operations \cite{KarapetyanIROS2019, KarapetyanFSR2019}. Depending on the type of the sensors and the aim of the exploration different patterns can be deployed. When 
a survey operation is performed with side-scan sonar the longitudinal coverage pattern, so-called L-cover method, is used to reduce the number of turns (see Figure \ref{fig:coverages}). Moreover, this approach can be enhanced by using implicit water current velocity information that the river meanders provide \cite{KarapetyanFSR2019}. When the resources are limited a partial coverage technique - Z-cover, can be use to collect the bathymetric data exploring the full width of the river with only one pass (see Figure \ref{fig:coverages}).

\section{Experimental Results}
The ASVs have been deployed in a variety of situations and with all the coverage patterns described above. In the next section we will present two coverage applications in a lake and a river environment. Then we will present results of controlling the ASV in the presence of external forces. 

\subsection{Coverage Applications}
\begin{figure*}[h]
		\centering
		\leavevmode
		\begin{tabular}{cc}
		   \includegraphics[width=0.47\textwidth]{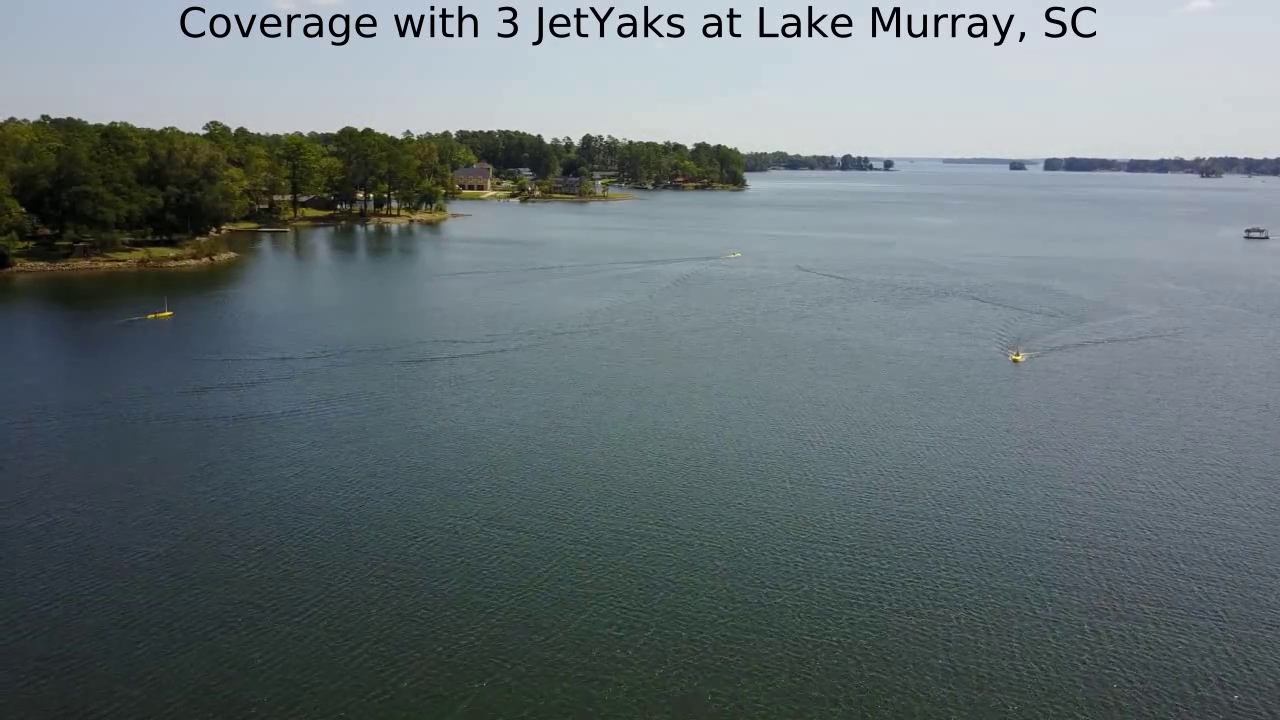}\label{fig:CL1}&
			\includegraphics[width=0.47\textwidth]{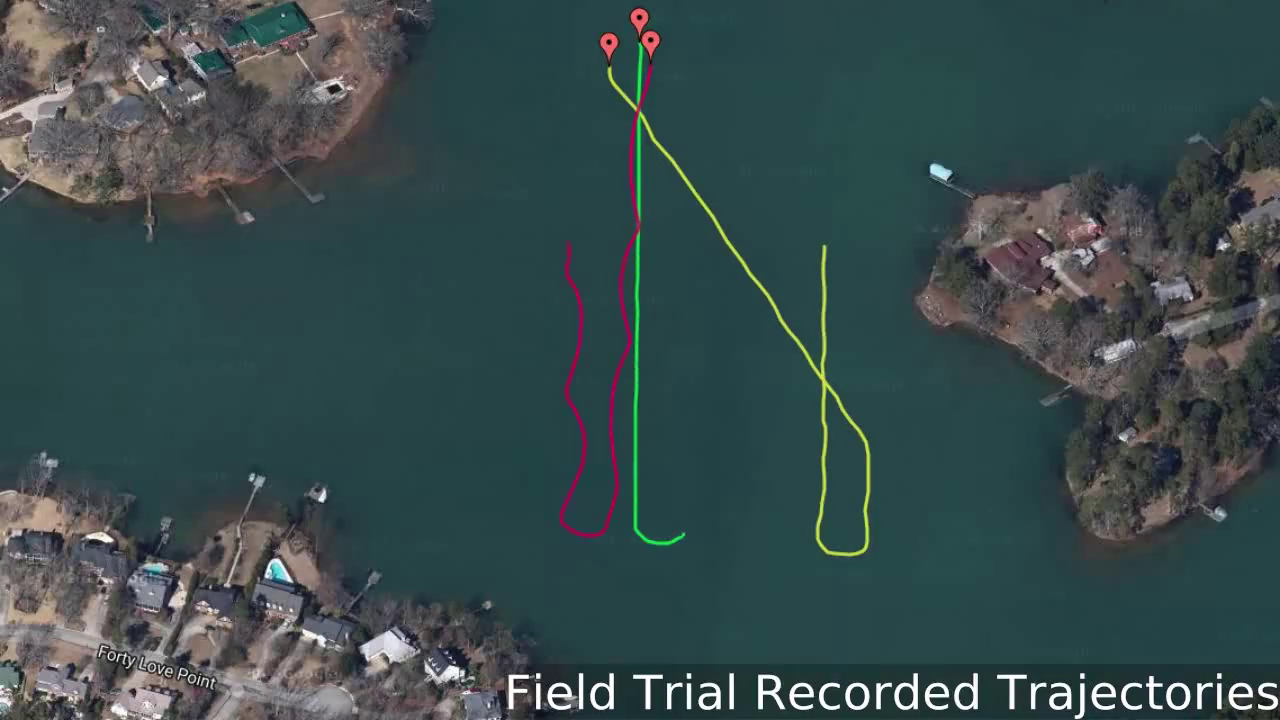}\label{fig:CL2}
		\end{tabular}
			\caption{Multi-ASV coverage at a lake: (a) Three ASVs operating at Lake Murray, SC; (b) Partial trajectories of the three ASVs during coverage; see \cite{karapetyan2018multi}.}
			\label{fig:CoverLake}
	\end{figure*}

Deploying multiple ASVs at a lake environment results in reduced operation times~\cite{karapetyan2018multi}. An optimization framework based on a TSP solver was used~\cite{LewisIROS2017} in addition to a partitioning of the resulting path to assign sub\hyp paths to individual ASVs~\cite{KarapetyanIROS2017}. Figure \ref{fig:CoverLake} presents three ASVs in action at Lake Murray, SC, USA, together with their partial trajectories. It is worth noting that even minimal disturbances produce deviations from a straight line path. This will result in overlapping coverage or leaving uncovered areas. Thus integrating the waypoint augmentation stage by providing feed-forward control based on the disturbances is an important addition for the proper operation of ASVs.

\subsection{Control Augmentation}

\begin{figure*}[h]
		\centering
		\leavevmode
		\begin{tabular}{cc}
			\subfigure[] {\includegraphics[width=0.450\textwidth]{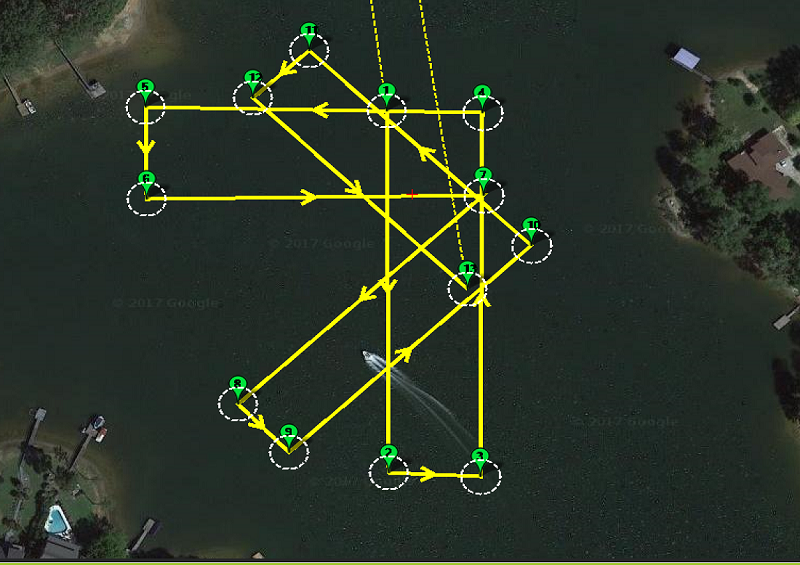}}\label{fig:lake_wp}&
			\subfigure[]{\includegraphics[width=0.450\textwidth]{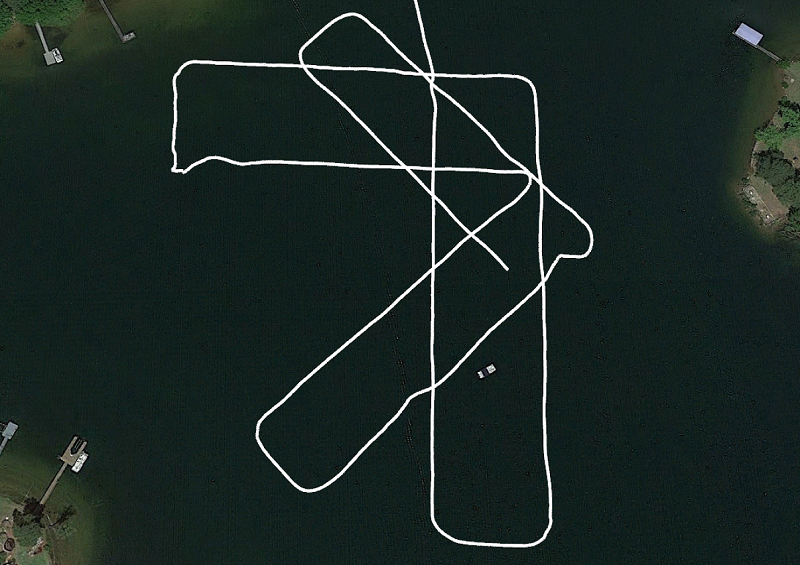}\label{fig:lake_actual}}
		\end{tabular}
			\caption{(a) One way-point mission with paths selected at 45 degree increments with respect to the predominant wind force. Target velocity for this iteration was 3 \SI{}{\meter\per\second}. (b) Actual path followed by the ASV on February 12, 2019 at Lake Murray, SC. }
			\label{fig:lake_test_patterns}
	\end{figure*}
	
\begin{figure*}[h]
		\centering
		\leavevmode
		\begin{tabular}{cc}
			\subfigure[]{\includegraphics[width=0.450\textwidth]{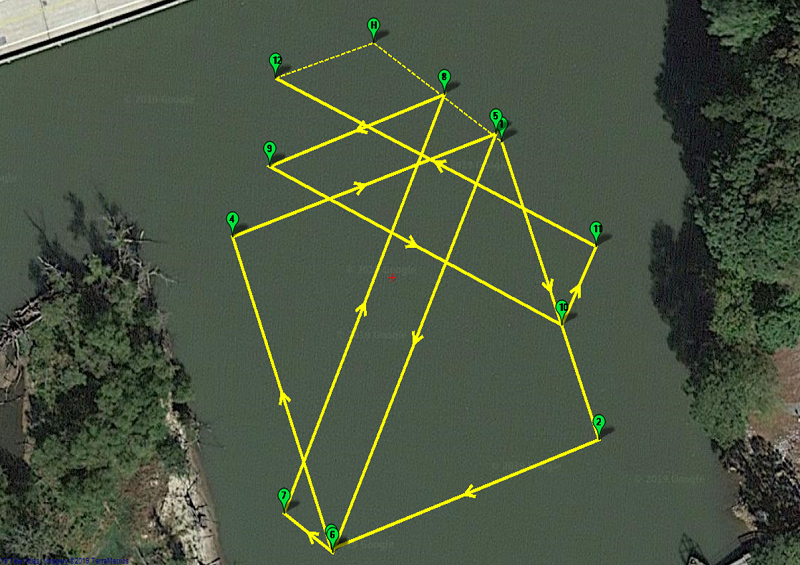}\label{fig:river_wp}}&
			\subfigure[]{\includegraphics[width=0.450\textwidth]{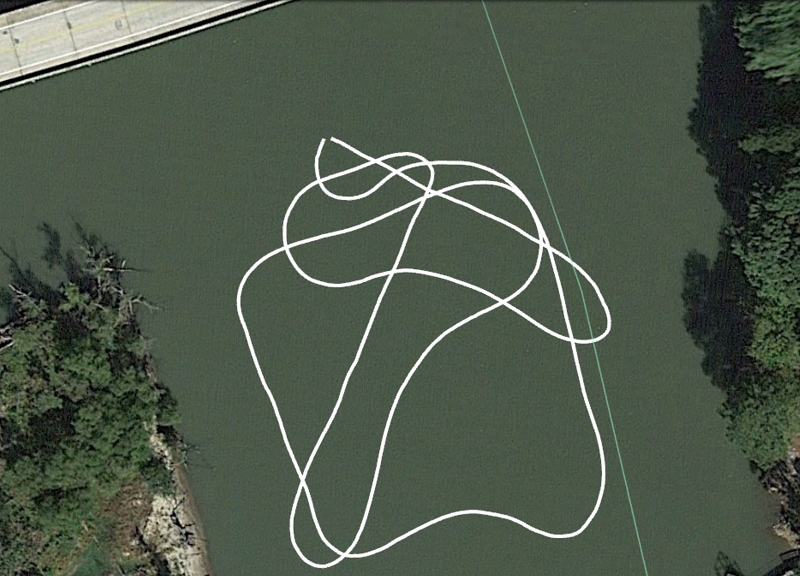}\label{fig:river_actual}}
		\end{tabular}
			\caption{(a) A Way-point mission with paths selected at 45 degree increments with respect to the predominant current force. (b) Actual path followed by the ASV on February 25, 2019 at Saluda River, SC. 
			\label{fig:river_test_patterns}}
	\end{figure*}

Figure \ref{fig:augmented_results} (a) shows the effect of the environmental forces on the planned trajectory when the speed of the ASV relative to the ground is increased. This resulted in too much error accumulation in the PID controller and made it unable to overcome the external forces. This will usually result in an overshoot scenario where the ASV harmonically oscillates back and forth over the desired trajectory. The latter is also illustrated in the experiments conducted on the Saluda River (see Figure \ref{fig:lake_test_patterns}) and Lake Murray (see Figure \ref{fig:river_test_patterns}). Note also that with the increase of the current speed  this behavior starts to present itself in trajectories perpendicular to the current.

When the augmentation stage is applied as can be observed from Figure \ref{fig:augmented_results} (b) a more precise path-following strategy has been achieved. These results serve as proof of concept for a feed-forward controller with waypoint augmentation that can improve the optimlaity of the coverage patters. As shown in Figure \ref{fig:augmented_results}, path following in currents in all orientations to the ASV is qualitatively improved. Qualitative results has been also reported in our previous work \cite{MoultonFSR2019} that show improvements in both maximum error and percentage of the path that is more than a meter from the target trajectory. Those results are confirming the intuition gained from qualitative results: the augmented control algorithm provides a better waypoint and thus path following strategy.

\section{Conclusions}

The paper presented a complete framework for an ASV to perform environmental monitoring using waypoint navigation. It combined four main components: modelling of the wind and water current forces for predictive control, predicting the effect of those forces on the displacement of the vehicle, augmentation of the waypoints, and different coverage techniques that generate waypoints. As qualitative results showed intermediate way\hyp point navigation resulted in smoother and more precise trajectories compared to the standalone on\hyp board PID controller (Figure \ref{fig:augmented_results}). We measured the max distance between the target trajectory and the executed one, which indicated that with the proposed approach we were able to reduce the error by almost 48$\%$. With this improvement the framework ensures more accurate coverage strategies. In addition, we are planning to perform more long distance experiments to calculate also the length of the path, to show the affect of the trajectory displacement.

\bibliographystyle{IEEEtran}
\bibliography{refs}
\nocite{*}

\end{document}